\documentclass[10pt, a4paper]{article}
\usepackage{lrec2022} 
\usepackage{multibib}
\newcites{languageresource}{Language Resources}
\usepackage{graphicx}
\usepackage{tabularx}
\usepackage{soul}
\usepackage{numprint}
\usepackage{multirow}
\usepackage{makecell}
\usepackage{enumerate}
\usepackage{footnote}
\usepackage[normalem]{ulem}
\usepackage{titlesec}
\titleformat{\section}{\normalfont\large\bfseries\center}{\thesection.}{1em}{}
\titleformat{\subsection}{\normalfont\SmallTitleFont\bfseries\raggedright}{\thesubsection.}{1em}{}
\titleformat{\subsubsection}{\normalfont\normalsize\bfseries\raggedright}{\thesubsubsection.}{1em}{}
\renewcommand\thesection{\arabic{section}}
\renewcommand\thesubsection{\thesection.\arabic{subsection}}
\renewcommand\thesubsubsection{\thesubsection.\arabic{subsubsection}}

\usepackage{epstopdf}
\usepackage[utf8]{inputenc}

\usepackage{hyperref}
\usepackage{xstring}
\usepackage{pgfplots}
\usepackage{color}

\newcommand{\corpus}{CoQAR}

\makeatletter
\newcommand\footnoteref[1]{\protected@xdef\@thefnmark{\ref{#1}}\@footnotemark}
\makeatother

\title{\corpus{}: Question Rewriting on CoQA}

\name{Quentin Brabant, Gwénolé Lecorvé, Lina M. Rojas-Barahona} 

\address{Orange Innovation \\
         2 Avenue Pierre Marzin. Lannion. France.\\
         \{quentin.brabant, gwenole.lecorve, linamaria.rojasbarahona\}@orange.com\\}

\abstract{
Questions asked by humans during a conversation often contain
contextual dependencies, i.e., explicit or implicit references to previous dialogue turns.
These dependencies take the form of coreferences (e.g., via pronoun use) or ellipses,
and can make the understanding difficult for automated systems.
One way to facilitate the understanding and subsequent treatments of a question is to rewrite it into an out-of-context form, i.e., a form that can be understood without the conversational context.
We propose \corpus{}, a corpus containing $4.5$K conversations from the Conversational Question-Answering dataset CoQA, for a total of $53$K follow-up question-answer pairs.
Each original question was manually annotated with at least 2 at most 3 out-of-context rewritings. 
\corpus{} can be used in the supervised learning of three tasks: question paraphrasing, question rewriting and conversational question answering.
In order to assess the quality of \corpus{}'s rewritings, we conduct several experiments consisting in training and evaluating models for these three tasks.
Our results support the idea that question rewriting can be used as a preprocessing step for question answering models, thereby increasing their performances.
\newline  }

\begin{document}

\maketitleabstract

\section{Introduction}
    Conversational Question Answering (CQA) \cite{reddy_coqa_2019,choi_quac_2018,saha2018complex} is a task in which a system interacts with a so-called \textit{student}. The interaction takes the form of a conversation, where the student asks questions, and the system is expected to provide the right answers.
    In this paper we focus on the case where the system searches for answers in a text passage,
    although settings relying on structured data (e.g. knowledge bases) also exist~\cite{saha2018complex}.
    Compared to non-conversational question answering (or QA for short), the system faces an additional difficulty: each question is asked in a \textit{conversational context} that consists in previous conversation turns; implicit references to the conversational context may happen in the form of ellipses and coreferences, making the understanding of questions more difficult for the system.
    
    One way to overcome this difficulty is Question Rewriting (QR), which consists in rewriting each original (\textit{in-context}) question into an \textit{out-of-context} question that is understandable by itself, i.e., that can be answered without knowing the conversational context.
    \newcite{vakulenko2021question} argue in favor of this approach by experimentally showing that adding QR as a pre-processing step of CQA models can improve their performances. They also claim that QR models offer several advantages, including the possibility of \textit{reuse}: a same QR model can be used as a preprocessing step for several existing (conversational or non-conversational) QA models and datasets. In particular, any existing non-conversational QA model (see, e.g., \cite{rajpurkar_know_2018,usbeck20188th}) can be immediately used for CQA. 
    
    In this paper, we present the \corpus{} corpus, which is an annotated subset of the CQA corpus CoQA~\cite{reddy_coqa_2019}. \corpus{} was obtained by asking specialised native speakers to annotate
    original questions with at least two and at most three distinct \textit{out-of-context} rewritings.
    Our contribution is two-fold.
    
    
    
    Firstly, we provide \corpus{}, which contains high-quality questions rewritings. The corpus is publicly available\footnote{\label{coqar}The COQAR dataset is publicly available at \url{https://github.com/Orange-OpenSource/COQAR}}; moreover, its annotations were conducted in accordance to ethical concerns: every annotator involved was properly hired.
    
   Secondly, we assess the quality of the annotations of \corpus{} through several experiments. We train Question Rewriting (QR) models, as well as Question Paraphrasing (QP) models on \corpus{} and other datasets. We then rate these models' outputs via human evaluation. We also evaluate QR models as preprocessing steps of (conversational and non-conversational) QA models. To this end, we compare the performance of a stat-of-the-art QA model with and without QR.
    
   Our results support the claim of \newcite{vakulenko2021question} that QR models can be successfully used in combination with existing QA models. Indeed, we found that adding QR as a preprocessing step boosts the performances of QA models and allows reusing non-conversational state-of-the-art QA systems while reducing performance degradation on CQA.


    
    In the remainder of this paper we present the related work in Section~\ref{sec:back}. We introduce \corpus{} in Section~\ref{sec:CoQAR}. We talk about the NLP task we use to evaluate the proposed annotations in Section~\ref{sec:nlp}. The evaluation and discussion are presented in Section~\ref{sec:eval} and Section~\ref{sec:disc},
    respectively.

\begin{table*}[t!]
            \centering
            \begin{tabular}{r|l}
                \Xhline{3\arrayrulewidth}
                passage & \makecell{This is the story of a young girl and her dog. The young girl and her dog set out\\ a trip into the woods one day. Upon entering the woods the girl and her dog found \\ that the woods were dark and cold [...].}\\
                \Xhline{3\arrayrulewidth}
            	question & What is the story about? \\
            	\hline
            	\multirow{3}{*}{rewrittings}
            	& What is the subject of this story? \\
            	& Who are the two main characters in the story? \\
            	& Who is this story centered on? \\
            	\hline
            	answer & A girl and a dog. \\
            	\hline
	            answer span & This is the story of a young girl and her dog. \\
            	\Xhline{3\arrayrulewidth}
            	question & What were they doing? \\
            	\hline
            	\multirow{3}{*}{rewrittings}
            	& What were the girl and her dog up to? \\
            	& What did the girl and her dog decide to do? \\
            	& What was the activity of the girl and the dog for the day? \\
            	\hline
            	answer & Set on on a trip \\
            	\hline
	            answer span & The young girl and her dog set out a trip \\
            	\Xhline{3\arrayrulewidth}
            	question & where? \\
            	\hline
            	\multirow{3}{*}{rewrittings}
            	& Where did the girl and her dog go on a trip? \\
            	& What location did the girl and her dog journey to? \\
            	& What place did the girl and her dog go on that day? \\
            	\hline
            	answer & the woods\\
            	\hline
	            answer span & set out a trip into the woods\\
            	\Xhline{3\arrayrulewidth}
            \end{tabular}
            \caption{Beginning of a passage from \corpus{} and of the corresponding conversation.}
            \label{tab:example}
        \end{table*}

\section{Related Work}\label{sec:back}
        CoQA \cite{reddy_coqa_2019} is a Conversational Question Answering dataset that was originally created for measuring the ability of machines to handle conversational question answering.
        It contains 8k conversations, which sum up to 127k questions with answers.
        Each dialogue was produced by two human annotators, one \textit{student} asking questions, and one \textit{teacher} providing answers.
        Each conversation is about a piece of text called \textit{passage}. 
        The questions are conversational, while each answer is provided in two forms: (1) the answer per-say, which is a short piece of text (not necessarily a full sentence); (2) the \textit{answer span}, which is a quote from the passage from which the answer is deduced.
        Many answers are a subsequence of the answer span; however, this is not always the case. For example, the answer to a yes/no questions is ``yes'' or ``no'', although those word usually do not appear in the answer span.
        Each passage belongs to one of seven domains; two of these domains only appear in the test set.
        Many questions require pragmatic reasoning, which makes CoQA a challenging evaluation dataset for conversational question answering systems.
        Moreover, the authors estimate that 70\% of the questions cannot be correctly understood without taking into account the context established during previous dialogue turns.
        Finally, some of those questions are not answerable based on the passage.
        The right answer to these question is represented by the special ``unknown'' string.

        Similar to our work, the corpus \textbf{}{CANARD} \cite{elgohary_can_2019} contains a subset of the corpus QuAC \cite{choi_quac_2018}, another dataset for CQA. As in CoQA, each QuAC dialogue was produced by two crowd workers (one student and one teacher) and answers are spans extracted from a given piece of text. However, on the contrary to CoQA, the student does not see the text from which answers are taken.
        As CoQA, it contains unanswerable questions.
        CANARD was created by manually annotating a subset of QuAC: each question in CANARD was associated to one single out-of-context rewriting.  The train/dev/test sets of CANARD respectively contain 5,571/3,418/31,538 questions.
        CANARD was used for evaluating the impact of QR on Question Answering models in \newcite{vakulenko2021question}.


\section{CoQA with Question Rewriting (\corpus{})}\label{sec:CoQAR}
        \corpus{}\footnoteref{coqar} was created from CoQA in a way that is analogous to how CANARD was created from QuAC.
        However, while CANARD was annotated using crowd-sourcing,
        we decided to hire two specialized native-speakers annotators.
        Their task was to annotate
        original (\textit{in-context}) questions from CoQA with at least two and at most three distinct \textit{out-of-context} rewritings.
        To make sure that they understand what was expected, we ourself annotated a dialogue and provided it as an example.
        An example of conversation annotated by the annotators is provided in Table \ref{tab:example}.

        While annotators were told to preserve the meaning of the original sentences, they were also asked to paraphrase in their rewritings. As a results, these annotations contrast with those of CANARD, where the structure of the original question is usually preserved in the rewriting.
        In total, $4.1k$ conversations of CoQA train set were annotated as well as all $500$ conversations of the dev set. Since the test set of CoQA is not available, no conversation were annotated from it. The train and dev sets of \corpus{} respectively contain 45k and 8k questions. Table \ref{tab:nb-rewritings} summarizes the number of questions that have 0,1,2 or 3 rewritings.

        \begin{table}
            \centering
            \begin{tabular}{cccccc}
                \hline
                & \multicolumn{4}{c}{Number of rewritings} &\\
            	& $0$ & $1$ & $2$ & $3$ & total \\
            	\hline
            	train & 365 & 108 & 31,378 & 13,210 & 45,061 \\
            	dev & 9 & 0 & 37 & 7,937 & 7,983 \\
            	\hline
            \end{tabular}
            \caption{Number of questions depending on the number of rewritings.}
            \label{tab:nb-rewritings}
        \end{table}
        
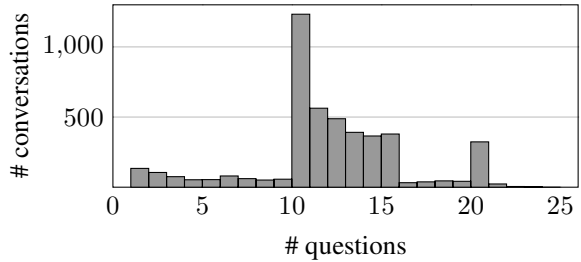
\begin{figure}[t]
            \centering
            \begin{tikzpicture}\centering
            \begin{axis}[
                ybar,
                xmajorgrids = false,
                height = 4cm,
                width = \columnwidth,
                major tick length = 0pt,
                minor tick length = 0pt,
                ymajorgrids = true,
                ymin = 0, ymax = 1300,
                xtick={0,5,...,30},
                ytick={500,1000},
                xmin = 0, xmax = 26,
                xlabel={\# questions},
                ylabel={\# conversations}
            ]
            \addplot+ [ybar interval, color=black!40, draw=black] coordinates { (1, 134) (2, 105) (3, 75) (4, 53) (5, 54) (6, 80) (7, 61) (8, 51) (9, 57) (10, 1233) (11, 563) (12, 488) (13, 391) (14, 365) (15, 379) (16, 32) (17, 38) (18, 45) (19, 42) (20, 323) (21, 23) (22, 5) (23, 3) (24, 0) (25, 1) };
            \end{axis}
            \end{tikzpicture}
            \caption{Distribution of conversations' length.}
            \label{fig:dialogue-length}
        \end{figure}

        Overall, passages contains from 75 to 1079 words, with an average of 275. Conversation length distribution is displayed in Figure \ref{fig:dialogue-length}.

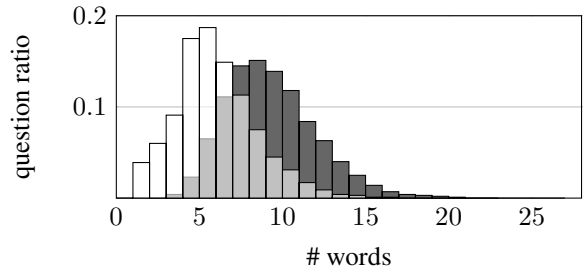
\begin{figure}[t]
            \centering
            \begin{tikzpicture}\centering
            \begin{axis}[
                ybar,
                xmajorgrids = false,
                height = 4cm,
                width = \columnwidth,
                major tick length = 0pt,
                minor tick length = 0pt,
                ymajorgrids = true,
                ymin = 0, ymax = 0.2,
                ytick={0.1,0.2},
                xtick={0,5,...,30},
                xmin = 0, xmax = 28,
                xlabel={\# words},
                ylabel={question ratio}
            ]
            \addplot+ [ybar interval, color=black!60, draw=black] coordinates{ (2, 0.0) (3, 0.004) (4, 0.023) (5, 0.065) (6, 0.111) (7, 0.145) (8, 0.151) (9, 0.139) (10, 0.118) (11, 0.084) (12, 0.063) (13, 0.04) (14, 0.025) (15, 0.014) (16, 0.007) (17, 0.004) (18, 0.003) (19, 0.002) (20, 0.001) (21, 0.0) (22, 0.0) (23, 0.0) (24, 0.0) (25, 0.0) (26, 0.0) (27, 0.0) };
            \addplot+ [ybar interval, fill=white, fill opacity=0.6, draw=black] coordinates { (0,0) (1, 0.039) (2, 0.06) (3, 0.091) (4, 0.175) (5, 0.187) (6, 0.149) (7, 0.113) (8, 0.075) (9, 0.045) (10, 0.031) (11, 0.017) (12, 0.009) (13, 0.004) (14, 0.003) (15, 0.001) (16, 0.001) (17, 0.0) (18, 0.0) (19, 0.0) (20, 0.0) (21, 0.0) (22, 0.0) (23, 0.0) };
            \end{axis}
            \end{tikzpicture}
            \caption{Distribution of length
            for original questions (white) and out-of-context rewritings (dark grey). Overlap of the distribution is light grey.}
            \label{fig:question-length}
        \end{figure}

        On average, out-of-context rewritings are longer (8.8 words) than the original questions (5.5 words); Figure \ref{fig:question-length} shows the question length distribution.

        Most conversations were annotated by only one annotator, but $50$ conversations were annotated by both.
        We relied on these conversations to analyse the annotations.
        We extracted two rewritings per question and per annotator and,
        using a pair of rewritings as references and the other as hypothesis, we computed the SacreBLEU score~\cite{post-2018-call} and the BERT-score~\cite{bert-score}. SacreBLEU gives us an insight on the similarity of the surface form of rewritings, while BERT-score gives us an insight on the semantic similarity.
        We obtained a SacreBLEU score of $32.67$ and a BERT-score of $90.22$: this suggests that the rewritings have diverse surface form while being close in terms of meaning.

\section{NLP Tasks}\label{sec:nlp}

    This section presents briefly the tasks of
    Question Paraphrasing (QP), Question Rewriting (QR) and
    Conversational Question Answering (CQA) that we used to evaluate the quality of the novel annotations of \corpus.
    
    \subsection{Question Paraphrasing (QP)}
    
        QP is the task of transforming a source question into a question with equivalent meaning but different surface form (syntax, lexicon, etc.). In this paper we consider the case where both the source and paraphrased questions are out-of-context questions. 
    	
        For each original question, \corpus{} provides several out-of-context rewritings.
        We can regard two out-of-context rewritings of a same original in-context question as the source and paraphrase questions in the QP task.
        
        We conducted experiments that consist in: (1) training QP models on \corpus{} and an additional dataset, namely Quora Question Pairs (QQP); (2) evaluating the paraphrases generated by the models, via the standard metrics BLEU and METEOR, as well as human evaluation. More details about the experiments are presented in Section~\ref{ssec:qp}.
   
    \subsection{Question Rewriting (QR)}
        
        In QR, the model receives as input an in-context question, its conversational context, and the associated passage. Its task is to generate an out-of-context rewriting of the question.
        
        We conducted the following experiment: (1) training QR models on \corpus{} and CANARD; (2) evaluating these models, via standard metrics and human evaluation as presented in Section~\ref{subs:qrew}. Furthermore, we evaluate these QR models on downstream conversational question answering as presented in the next section and in Section~\ref{ssec:evalcqa}.


    \subsection{Conversational Question Answering (CQA)}
        We consider CQA as a task for indirectly evaluating QR models.
        Typically, the inputs to a CQA neural model are:
        a question,
        its conversational context (i.e. the sequence of previous questions and answers),
        and the associated passage.
        


        A challenge for conversational question answering was also released with CoQA\footnote{\url{https://stanfordnlp.github.io/coqa/}}.  The models are evaluated with the F1 score~\cite{reddy_coqa_2019}. Transformers have been successfully used in this task: to the time this paper was written, the best model (a RoBERTa-based model~\cite{ju2019technical}) got $90.7$ of overall F1 measure, overcoming human performance $88.8$. 
        
        Our goal is to indirectly assess the quality of QR by comparing the performance of a model taking original questions and their context as inputs with a model using out-of-context rewritings instead. In other words, we would like to know whether replacing the original question with its conversational context by the out-of-context rewriting has a positive impact on answer extraction. First, we evaluate the impact of rewritten questions in the performance of a RoBERTa baseline~\cite{liu2019roberta}.
        Second, in order to assess the reusability of QR models trained on \corpus{}, we further evaluate a state-of-the-art non-conversational QA model trained on SQuAD~\cite{rajpurkar_know_2018} by testing it with the rewritten questions. Please refer to Section~\ref{ssec:evalcqa} for more details about the evaluation of QR for this task.

\section{Evaluation}\label{sec:eval}
    In this section we present the settings and results of our experiments.
    Those involve the fine-tuning of the T5 and BART pretrained transformer models,
    with various training sets.
    We refer to fine tuned models with names of the form: ``model(training-data-source)''. For example, T5(\corpus{}) will refer to a T5 model that was fine-tuned on data from \corpus{}.
    \subsection{Question Paraphrasing}\label{ssec:qp}
        We first train QP models on \corpus{} and Quora Question Pairs (QQP), then we evaluate the quality of the paraphrases generated by the models in terms of BLEU, METEOR and human evaluation. 
    
        \paragraph{Datasets:} Each QP model was trained on a set of pairs consisting of a source question and its paraphrase, which are both out-of-context questions. We extracted such pairs from \corpus{} and QQP. Since original questions of \corpus{} have several out-of-context rewritings, we built pairs by associating rewritings of a same original question. This corresponds to a total of $237$K paraphrase pairs, for an average of $1.9$ paraphrase per out-of-context question.
        The QQP corpus is not a QA corpus: it was originally proposed as a Kaggle challenge to detect duplicate questions from Quora, a collaborative QA website where users can post their own questions or reply to those asked by others. The QQP corpus is composed of 404K question pairs, out of which 37\,\% are flagged as duplicates. We regard duplicate questions as paraphrases;
        assuming the transitivity of the semantic equivalence relation, clusters of paraphrases can be built.
        This results in a total of 710K paraphrase pairs, where each question is linked to 4.8 paraphrases on average. Clusters are partitioned into a training and test set with ratios 80 and 20\,\%, respectively.

        \paragraph{Models:} Three QP models are built by fine-tuning a pretrained BART model~\cite{lewis_bart_2020}  (\textit{base} version\footnote{\url{https://huggingface.co/facebook/bart-base}}) on paraphrased question pairs.
        Each model is trained on one of three set of pairs: (1) pairs coming from \corpus{}, (2) pairs coming from QQP, (3) pairs coming from both QQP and \corpus{}. 
        The models are fine-tuned during 2 epochs with batches of 10 samples. Optimization is done using AdamW, and static learning rate $5\times10^{-5}$.
        %
        \\
        \textit{Remark: Experiments with T5 models were also carried out but leading to slightly worse results. Thus, they are not reported here.}
     
        \paragraph{Objective Evaluation:} Table~\ref{tab:meteor_paraphrases} compares the BLEU and METEOR scores obtained by the fine-tuned BART models against a naive model that copies the input sentence as output. BLEU is provided for comparison purposes, even though it is known as less relevant for this task. Scores are measured on the test set of \corpus{} and QQP.
        %
                
            
                

        \begin{table}
            \centering
            \setlength{\tabcolsep}{3.5pt}
            \begin{tabular}{llcc}
                \hline
                Test set & Model & BLEU & METEOR \\
                \hline
                \multirow{4}{*}{\corpus{}} & Naive (copy)          &  0.694 & 0.492 \\
                                           & BART(\corpus{})     &  0.705 & \bf{0.537} \\
                                           & BART(QQP)            &  0.673 & 0.464 \\
                                           & BART(\corpus{}+QQP)  &  \bf{0.737} & 0.526 \\
                \hline
                \multirow{4}{*}{QQP} & Naive (copy)           & \bf{0.737} & \bf{0.634} \\
                                     & BART(\corpus{})       &  0.626 & 0.445 \\
                                     & BART(QQP)             & 0.695 & 0.619 \\
                                     & BART(\corpus{}+QQP)   & 0.692 & 0.611 \\
                \hline
            \end{tabular}
            \caption{BLEU and METEOR scores on the test of \corpus{} and QQP for various models. The Naive model simply outputs its input without any modification.}
            \label{tab:meteor_paraphrases}
        \end{table}
        
        \begin{table*}[h!]
            \centering
            \begin{tabular}{llcccc}
                \hline
                \multirow{2}{*}{Test set} & \multirow{2}{*}{Model} & \multicolumn{2}{c}{Meaning preservation} & \multicolumn{2}{c}{Linguistic correctness} \\
                &  & {MOS} & {(Std dev.)} & {MOS} & {(Std dev.)}\\
                \hline
                \multirow{2}{1cm}{\corpus{}} & Reference      & 3.82 & (1.04) & 4.46 & (0.83) \\
                                             & BART(\corpus{})      & \textbf{3.97} & (1.15) & \textbf{4.54} & (0.75) \\
                \hline
                                             
                                             & Reference      & 3.32 & (1.28) & 4.33 & (1.01) \\
                QQP                          & BART(QQP)            & \textbf{3.64} & (1.12) & 4.37 & (0.91) \\
                                             & BART(\corpus{}+QQP)  & \textbf{3.65} & (1.21) & \textbf{4.51} & (0.66) \\
                \hline
                CANARD                          & BART(\corpus{})  & \textbf{4.15} & (1.04) & \textbf{4.41} & (0.94) \\
                \hline
            \end{tabular}
            
            \caption{
            Results of the human evaluation of QP.}
            \label{tab:human_eval_paraphrases}
        \end{table*}
        
        First, the results show high values for the naive approach. This indicates (not surprisingly) that the source questions and their paraphrases are lexically close, especially in QQP.
	    On \corpus{}'s test set, BART models whose training incorporates \corpus{} data perform better than the naive model, demonstrating that fine-tuning enabled models to learn the task; on the other hand, on QQP's test set, the naive model gives the best results.
	    These observations suggest that QQP may not be relevant for training and evaluating paraphrase generation models.
	    Finally, we observe that using crossed data (training on \corpus{} and testing on QQP, and vice versa) results quite logically in a loss of performance.
            
        \paragraph{Human Evaluation:} Two Mean Opinion Score (MOS) evaluations were carried out on 12 human testers who were asked to judge the quality of paraphrases.
        The objective is to complete observations from the automatic evaluations, as well as to study how \corpus{} can benefit to the task on other datasets.
        We considered three corpora: \corpus{}, QQP and CANARD.
        For each corpus, 50 source questions were randomly selected, and were paired with several paraphrases:
        \begin{itemize}
            \item one paraphrase from the corpus, to which we refer as the \textit{reference};
            \item one or several paraphrases generated by different BART models: each source question from \corpus{} and CANARD is paired with a paraphrase generated by BART(\corpus{}), while each source question from QQP is paired with one paraphrase generated by BART(QQP) and one generated by BART(\corpus{}+QQP).
        \end{itemize}
        %
        In a first evaluation phase, testers were asked to judge the semantic similarity between two questions presented to them. Their opinion could be given on a 5-point scale: (1) ``totally different"; (2) ``mostly different"; (3) ``half similar/half different"; (4) ``mostly similar"; (5) ``perfectly similar".
        In the second evaluation, each tester rated the linguistic correctness of single questions, independently of their meaning, on a similar scale to that used for semantic similarity.
        Each question pair (meaning preservation experiment) and single sentence (linguistic correctness) received 2 ratings.

        Table \ref{tab:human_eval_paraphrases} reports average values and standard deviation obtained for each MOS test.
        The main conclusions are given below, along with $p$-values from Mann-Withney U tests when relevant to assess the statistical significance between to mean values\footnote{As a reminder, the $p$-value measures the probability that the difference between two values is due to chance.}.
        
        On \corpus{}, paraphrases generated by BART obtain higher mean scores than the references, although the observed difference might be due to chance, both for meaning preservation ($p=0.069$) and linguistic correctness ($p=0.4$). This confirms that fine-tuning has indeed enabled the model to learn the task, as suggested by the BLEU and METEOR scores.
    	On QQP also, BART models generalize well as they exceed references in terms of meaning preservation, although the difference might again be due to chance ($p=0.091$). Adding \corpus{} to the train set does not improve meaning preservation, and the slight increase in linguistic correctness is not statistically significant ($p=0.37$).
    	When comparing the second and last line of the table, it seems that the BART model learned on \corpus{} transfers well to CANARD. However, it is possible that rewritings from CANARD are easier to paraphrase than those from \corpus{}.
    	Finally, is worth noting that QQP reference paraphrases obtained lower average meaning preservation scores than \corpus{} paraphrases ($p=0.019$).
    	A manual investigation in QQP indeed shows that some questions are linked to more (or less) generic ones: for instance, ``Given that \textit{C}, what is \textit{A}?" redirected to ``What is \textit{A}?", or ``What is \textit{A}?" redirected to ``What are \textit{A} and \textit{B}?"). While this makes sense for helping users finding answers, these questions are not semantically equivalent.
    	These observations suggest that QQP may not be relevant for training and evaluating paraphrase generation models.

        Overall, the experiments demonstrate that \corpus{} is conclusive to perform paraphrase generation on questions.

    \subsection{Question Rewriting}
    \label{subs:qrew}
           \begin{table}[ht]
            \centering
            \setlength{\tabcolsep}{1.5pt}
            \begin{tabular}{@{~}llcc@{~}}
                \hline
                Test set & Model / train set & BLEU & METEOR \\
                \hline
                \multirow{3}{*}{\corpus{}} & T5(\corpus{})      &  0.38 & 0.58 \\
                                           & T5(CANARD)     & 0.32  &  0.53 \\
                                           & T5(\corpus{}+CANARD)  & \textbf{0.39} & \textbf{0.59} \\
                \hline
                \multirow{3}{*}{CANARD} & T5(\corpus{})    & 0.31  & 0.57 \\
                                     & T5(CANARD)          & \textbf{0.47}  & \textbf{0.69} \\
                                     & T5(\corpus{}+CANARD)  & 0.44 & 0.66 \\
                \hline
            \end{tabular}
            \caption{BLEU and METEOR scores obtained by the Question Rewriting models.}
            \label{tab:scores-rewriting}
        \end{table}   
        \paragraph{Datasets.}
        For training and evaluation, we rely on CANARD and \corpus{}.
        For CANARD, we use the original train/dev/test splits.
        For CoQAR, we use the original dev set as test set, and split the original train set into a train set and dev set, in such manner that CANARD and CoQAR dev sets have the same size.
        For training, we also make use of a mixture of CANARD and \corpus{}, that we refer to as \corpus{}+CANARD, whose train and dev sets are, respectively, the union of both corpora's train and dev sets.
        We train three variants of the QR model: one variant is trained on CANARD, one is trained on \corpus{}, and the third one is trained on a mixture of both datasets.
           \begin{table*}[h!]
            \centering
            \begin{tabular}{llcccc}
                \hline
                \multirow{2}{*}{Test set} & \multirow{2}{*}{Model} & \multicolumn{2}{c}{Meaning preservation} & \multicolumn{2}{c}{Linguistic correctness} \\
                
                & & {MOS} & {(Std dev.)} & {MOS} & {(Std dev.)}\\
                \hline
                \multirow{2}{1cm}{\corpus{}} & Human rewriting      & \textbf{4.5} & (0.86) & \textbf{4.86} & (0.45) \\
                                             & T5(\corpus{})      & 3.82 & (1.42) & 4.66 & (0.82) \\
                \hline
                                             
                                             & Human rewriting      & \textbf{4.60} & (0.96) & 4.7 & (0.89) \\
                CANARD                          & T5(CANARD)           & 3.92 & (1.34) & 4.43 & (1.08) \\
                                             & T5(\corpus{}+CANARD)  & 3.96 & (1.47) & \textbf{4.76}  & (0.77)  \\
                \hline
            \end{tabular}
            
            \caption{Results of the human evaluation of QR.}
            \label{tab:human_eval_rewriting}
        \end{table*}      
        
        \paragraph{Model:} We train a QR model based on T5 on three datasets: \corpus, CANARD, and \corpus{}+CANARD.
        For each dataset, we fine-tune the small 1.1 version of T5\footnote{\url{https://huggingface.co/google/t5-v1_1-small}}.
        We use AdamW optimizer, with initial learning rate $5\times10^{-5}$ and no weight decay.
        After each epoch, the model is evaluated on the dev set using METEOR.
        We stop training as soon as the last obtained METEOR score is smaller than the two previous ones; we then keep the model that yielded the highest score.\\
        \textit{Remark: BART models were also trained on the QR task; their BLEU and METEOR scores were overall similar but slightly worse than those of T5 models, thus we excluded them from the human evaluation phase and omitted them from the reported results.}

        \paragraph{Objective Evaluation}

        Table~\ref{tab:scores-rewriting} compares BLEU and METEOR scores obtained by the three fine-tuned T5 models. Scores are measured on \corpus{} and CANARD test sets.
        Not surprisingly, performance drops when the models are tested on a data source which differs from the training data source (2st and 4th rows). On the contrary, mixing both corpora during training results in a unique model that performs well on both test sets (3rd and 6th rows).
        Scores are higher when testing on CANARD: this is again not surprising, since \corpus{} rewritten questions have more diverse surface forms than those in CANARD, which are more similar to the original questions.

        \paragraph{Human Evaluation}
        Two Mean Opinion Score (MOS) evaluations were carried out on 8 human testers who were asked to judge the quality of rewritten questions.
        We sampled 50 original questions from \corpus{} and 50 original questions from CANARD.
        Each original question was then paired with several rewritings:
        \begin{itemize}
            \item one rewriting from the corpus, to which we refer as the \textit{reference};
            \item one or several rewritings generated by different T5 models: each source question from \corpus{} is paired with a rewriting generated by T5(\corpus{}), while each source question from CANARD is paired with one rewriting generated by T5(CANARD) and one rewriting generated by T5(\corpus{}+CANARD).
        \end{itemize}
        The pairs were then used in two evaluations.
        
        In the first evaluation, rewritten questions were presented to human testers, together with the original question and its context (preceding dialogue turns and the corresponding text passage). Testers assessed the semantic similarity of the rewritten and original questions.
        In the second evaluation, rewritten questions were presented alone to the testers for them to assess linguistic correctness.
        Both semantic similarity and linguistic correctness were evaluated on the 5-points scale introduced in \ref{ssec:qp}.
        In the end, each rewritten question received one rating for semantic similarity and one for linguistic correctness.
        The results are reported in Table~\ref{tab:human_eval_rewriting}.

        We see that QR models obtain scores that are clearly below human performance in terms of meaning preservation. We also observe that the T5 model that was trained on \corpus{} and CANARD obtains higher linguistic correctness scores than the model that was only trained on CANARD, and this result does not seem due to chance (a Mann-Whitney U test gives a $p$-value of $0.026$). It is plausible that, although adding data from \corpus{} to the training set does not improve meaning preservation, it improves linguistic correctness because of its greater diversity in term of rewritings' surface forms.
        Finally, note that the scores in Table \ref{tab:human_eval_rewriting} should not be compared with those of Table \ref{tab:human_eval_paraphrases}, because the sets of testers only partially overlap.
        
     \subsection{Conversational Question Answering}  
     \label{ssec:evalcqa}
     We would like to assess the impact of QR on state-of-the art models for CQA by answering the following question: would the models be able to extract the correct answer from the passage without dealing with the conversational context? To this aim we propose three experiments in which we train and evaluate a transformer on several variations of QR: no rewriting, human rewriting and model rewriting.
             \begin{table}[t]
            \centering
            \setlength{\tabcolsep}{10pt}
            \begin{tabular}{lcc}\hline
                 QR mechanism&F1&EM  \\\hline
                 None (question+context)&\bf{68.13}&\bf{49.63} \\
                 Human rewriting&63.26&45.10 \\ 
                 T5(\corpus{}+CANARD)&63.30&44.97\\
                 
                 \hline
            \end{tabular}
            \caption{Results of the CQA evaluation.}
            \label{tab:convqares}
        \end{table}   
     \paragraph{Datasets.} We use \corpus{}, with distinct rewriting.
     \begin{enumerate}[i]
         \item No rewriting: the orginal dataset, taking into account the conversational context.
         \item Human rewriting:  the dataset containing only the question rewritten by human annotators, ignoring completely the conversational context.
         \item QR model: instead of using human annotations we use questions that were generated automatically by the T5(\corpus{}+CANARD) model presented in Section~\ref{subs:qrew}.
     \end{enumerate}
        \begin{table*}[t]
            \centering
            \setlength{\tabcolsep}{10pt}
            \begin{tabular}{llccc}\hline
                 Test set&QR mechanism&F1&EM & unknown  \\\hline
                 \multirow{5}{*}{\corpus{}}&None (question+context) &35.89&8.22 & 0.5\\
                 &Human rewriting&\bf{42.21}&\textbf{9.23} & 0.4 \\ 
                 &T5(\corpus{})&41.56 &9.09 & 0.4\\
                 &T5(CANARD)&39.84&8.84 & 0.4\\
                 &T5(\corpus{}+CANARD)&41.80&9.17 & 0.4\\\hline
                 \multirow{5}{*}{CANARD}&None (question+context) &27.46&\bf{17.63} & 16.9\\
                 &Human rewriting&\textbf{28.02}&16.08 & 15.3\\ 
                 &T5(CANARD)&27.21&15.76 & 15.7 \\
                 &T5(\corpus{})&27.06&16.44 &  14.9\\
                 &T5(\corpus{}+CANARD)&27.18&16.03 & 15.3\\\hline
            \end{tabular}
            \caption{Results of the reusability evaluation.
            The ``unknown'' column contains the percentage of questions with no answer where an exact match is obtained. }
            \label{tab:qares}
        \end{table*}       
     \paragraph{Model.} For the CQA experiments, we train and evaluate a RoBERTa\footnote{\url{https://huggingface.co/}} transformer on \corpus{} with the distinct rewriting mechanisms described above. We fine tune the model during up to 5 epochs. We used Adam optimiser~\cite{kingma_adam_2017}, with learning rate of $5e-5$ and 12 gradient accumulation steps.
     

        
    \paragraph{CQA evaluation.}
    Results are presented in Table~\ref{tab:convqares}.
    Surprisingly, resolving the context with human question rewriting does not seem to help RoBERTa to better identify the answer in terms of F1 and exact match (EM) as defined in~\cite{rajpurkar2016squad}. We obtained an  F1 and EM gain of $4.87$ and  $4.53$ respectively of the original in-context questions over the out-of-context human rewritings.
   
   Unlike ~\newcite{vakulenko2021question}, where results of the same task are reported on CANARD, the setting relying on original questions (referred to as CANARD\_O)
    and the one relying on human-written questions (CANARD\_H)
    respectively obtain $53.65$ and $57.12$ F1 scores,
    which correspond to a gain of $3.47$ points for human rewriting.
    We suspect that the self-attention mechanism of RoBERTa solves the coreferences and ellipsis present in short in-context questions limited by the separation token from the context and the passage. While processing a long self-contained rewriting might be more difficult.  These results confirm the good performance of RoBERTa  on the original task of CQA \cite{ju2019technical}.


    Interestingly, automatically rewritten questions trained on both \corpus{} and CANARD obtained similar performance than human rewritings, although human rewriting, got a slightly better EM.  These results are comparable with the ones reported on CANARD in~\newcite{vakulenko2021question}. 
    %
        %

    \paragraph{Reusability evaluation.}
    To assess the reusability of the QR models trained on \corpus{}, we compare the performances, on \corpus{} and CANARD, of an existing QA model, with several question rewriting techniques, including QR.
    The considered QA model is the hugging-face \textit{distilbert-base-uncased-distilled-squad}
, which was trained on SQuAD.
    We adopted the same preprocessing as before: (i) no rewriting, (ii) human rewriting, (iii) QR model.
        %

    Table~\ref{tab:qares} shows that DistilBERT obtains higher F1 scores on \corpus{}.
    For both test set, the best F1 scores are obtained when using human-rewritten questions.
    In terms of exact match, better results are obtained on CANARD: however, almost all exact matches are obtained on questions whose answer is ``unknown''. This could be explained by the fact that questions with unknown answers constitute about $18\%$ of questions in CANARD, but less than $2\%$ in \corpus{}.
    Overall, it seems that the chosen QA model cannot handle CANARD correctly, independently on the QR step.
    On \corpus{}, using human rewritings yields a significant increase of F1-score: from $35.90$ F1 to $42.21$.
    Interestingly, QR models produce results that come very close to human rewriting.
    Thus, the results on \corpus{} suggest that the QR models are able, as a pre-processing step, to improve the results of simple QA systems on CQA.


\section{Conclusion}
\label{sec:disc}
In this paper, we presented \corpus{}, a subset of CoQA where questions were annotated with out-of-context paraphrases.
We took ethical concerns seriously, thus we hired two specialised native annotators for the task. 
Each question was annotated with several parapharses, and
we demonstrated the richness of these paraphrases in terms of diversity in the surface form.
Moreover, we evaluated the quality of the annotations via three tasks: QP, QR and CQA.

The results of the QP experiments suggest that \corpus{} is more adapted to the task of Question Paraphrasing than QQP.
Moreover, the human evaluation in Subsection \ref{subs:qrew} shows that the out-of-context rewritings of \corpus{} are approximately as good as those of CANARD in terms of linguistic correctness and semantic similarity. 
This conclusion is also supported by the results of our experiments on QP and QR, where adding data from \corpus{} to QQP or to CANARD during training does improve linguistic correctness.

Finally, although the results of our experiments confirm that QR performed either by humans or by models, does not improve the performance of CQA; it does enable the usage of non-conversational QA in CQA settings.


\section{Acknowledgments}
The work of recruiting and coordinating annotators was done by ELDA\footnote{\url{http://www.elra.info/en/about/elda/}}. In particular, we would like to thank Khalid Choukri, Lucille Blanchard, and Marwa Hadj Salah. This work was fully funded by Orange Innovation. We thanks all the member of the DATA-AI department for their support. This work was granted access to the HPC resources of IDRIS under the allocation 2021-AD011011407 made by GENCI 

\section{Bibliographical References}

    \bibliographystyle{lrec2022-bib}
    \bibliography{bib}

\end{document}